\documentclass{article}

\usepackage[nonatbib, final]{nips_2018}

\usepackage[natbib=true,backend=biber,
style=numeric-comp,sorting=none,block=none,
giveninits=true,doi=false,isbn=false,date=year,
mincitenames=1,maxcitenames=2, 
maxbibnames=3,minbibnames=1,   
bibencoding=utf8]{biblatex}
\renewbibmacro{in:}{}
\AtEveryBibitem{\clearlist{language}}
\usepackage[utf8]{inputenc} 
\usepackage[T1]{fontenc}    
\usepackage{hyperref}       
\usepackage{url}            
\usepackage{booktabs}       
\usepackage{amsfonts}       
\usepackage{nicefrac}       
\usepackage{microtype}      
\usepackage{subcaption}
\usepackage{graphicx}
\usepackage{bm}             
\usepackage{algorithm}      
\usepackage[noend]{algpseudocode} 
\usepackage{enumitem}

\title{Coupling weak and strong supervision for classification of prostate cancer histopathology images}

\author{
  Eirini Arvaniti \\
  ETH Zurich\\
  \texttt{arvaniti@imsb.biol.ethz.ch} \\
\And
 Manfred Claassen \\
 ETH Zurich\\
 \texttt{claassen@imsb.biol.ethz.ch} \\
}

\begin{document}

\maketitle

\begin{abstract}
  Automated grading of prostate cancer histopathology images is a challenging task, with one key challenge being the scarcity of annotations down to the level of regions of interest (\textit{strong labels}), as typically the prostate cancer Gleason score is known only for entire tissue slides (\textit{weak labels}). In this study, we focus on automated Gleason score assignment of prostate cancer whole-slide images on the basis of a large weakly-labeled dataset and a smaller strongly-labeled one. We efficiently leverage information from both label sources by jointly training a classifier on the two datasets and by introducing a gradient update scheme that assigns different relative importances to each training example, as a means of self-controlling the weak supervision signal. Our approach achieves superior performance when compared with standard Gleason scoring methods.
\end{abstract}

\section{Introduction}

Therapeutic decisions for prostate cancer, the second most common cancer in men~\cite{Cancer_Genome_Atlas_Research_Network2015-aw}, are to a large extent determined by the Gleason score~\cite{Gleason1992-ji, Epstein2016-cj}. The \textit{Gleason score}, assigned by pathologists on the basis of microscopic examination of patient tissue slides, is computed as the sum of the primary and secondary \textit{Gleason patterns} observed in the tissue. \textit{Gleason patterns} are local prostate gland formations characteristic of different cancer grades. They are assigned a numeric value in the range $3 - 5$ indicating the grade, with 5 being the highest grade. The \textit{Gleason score} is the sum of the two most prevalent \textit{Gleason patterns}, and thus takes a numeric value in the range $6 - 10$. Gleason score assignment is a time-consuming and challenging task even for domain experts, with high inter-pathologist variability rates~\cite{Gleason1992-ji, Salmo2015-jj, Arvaniti2018-fw} due to the need to examine large and heterogeneous areas of tissue. Thus, an automated decision-support solution would be highly desirable.

One of the main obstacles in designing such a solution is the scarcity of high-quality pixel- or ROI-level \textit{Gleason pattern} annotations, as typically only the overall \textit{Gleason score} is reported for each patient, e.g. Gleason 6=3+3, Gleason 7=4+3 etc. (a.k.a. \textit{weak labels}). Recent work on whole-slide image (WSI) Gleason score classification has focused on standard supervised learning~\cite{Kallen2016-nv, Del_Toro2017-gk} and unsupervised domain adaptation~\cite{Ren2018-rg}, always on the basis of Gleason score weak labels available from either The Cancer Genome Atlas (TCGA)~\cite{Cancer_Genome_Atlas_Research_Network2015-aw} or in-house private datasets. Recently, \textcite{Arvaniti2018-fw} published a Tissue Microarray (TMA) dataset with ROI-level \textit{Gleason pattern} annotations, which is a complementary resource providing a smaller but strongly-labeled dataset. In this study, we focus on classification of TCGA whole-slide images into low ($\leq 7$) and high ($\geq 8$) Gleason score classes on the basis of the TCGA and TMA datasets, and propose a new approach that efficently leverages information from the two sources, combining weak and strong labels. Our approach is similar in nature to \cite{Dehghani2017-pq}, where a confidence network was used to provide weights for controlling the weak supervision signal. Here, we do not have access to a confidence score-generating mechanism and, as an alternative, propose to obtain per-example confidence weights from the target network itself.

\section{Methods}

\textbf{Problem formulation.} Let $S$ denote a strongly-labeled dataset with examples $(x_s, y_s)$ and $W$ a weakly-labeled dataset with examples $(x_w, y_w)$. In our case, $(x_s, y_s)$ correspond to TMA image patches with local Gleason annotations and $(x_w, y_w)$ correspond to WSI image patches that adopt the Gleason label of the entire WSI. The task is to build a classifier on the categories $C$ in $W$, using the data in $W \! \cup \! S$. The challenges associated with this task are that (a) the images in $S$ and $W$ do not follow the same distribution and (b) the labels in $S$ and $W$ describe the corresponding images at a different level (locally in $S$ vs globally in $W$).

\textbf{Addressing the covariate shift.} A classifier is likely to benefit from additional training data when this new data follows the distribution of the domain of interest. However, the TMA and TCGA datasets were generated by different institutions, which implies possible differences in tissue preparation, staining and digitization. As a first step, we investigated which data transformations/training strategies are necessary for obtaining good predictions on the TCGA dataset (\textit{target domain}), if we \textit{exclusively} use the TMA labels (\textit{source domain}) for training. We considered the following approaches and applied them incrementally (e.g. stain transfer also performs color augmentation etc.):

\begin{itemize}[leftmargin=0.5cm, itemsep=0.1pt, topsep=0pt]
\item \textbf{\bm{$S-$}only w/o color augmentation.} Trains a classifier using exclusively data from $S$.

\item \textbf{\bm{$S-$}only w/ color augmentation.} Performs random image color perturbations prior to training.

\item \textbf{\bm{$S-$}only w/ stain transfer.} For each source domain example, randomly selects an example from the target domain and transfers its stain colors~\cite{Vahadane2016-fs} to the source domain example.

\item \textbf{Symmetric domain adaptation.} Obtains a domain-invariant classifier by jointly training on the source and target domain. We considered methods that encourage domain-invariant features via MMD minimization ~\cite{Tzeng2014-bm}, feature covariance alignment (CORAL)~\cite{Sun2016-jy} and domain adversarial training~\cite{Ganin2016-gn}.

\end{itemize}

\textbf{Combining weak and strong supervision.} As a second step, and assuming we have already selected a strategy for reducing the covariate shift, we investigated how to best integrate the two available data sources. The following approaches were considered:

\begin{itemize}[leftmargin=0.5cm, itemsep=0.1pt, topsep=0pt]
\item \textbf{\bm{$W\!-$}only, \bm{$W \! \cup \! S$}.} Supervised learning using exclusively weakly-labeled data ($W\!-$only), or both weakly- and strongly-labeled data ($W \! \cup \! S$).

\item \textbf{MIL-based Weak Supervision (MIL-WS).} In the spirit of recent multiple instance learning (MIL)-based approaches in computational pathology~\cite{Hou2016-ws, Campanella2018-dt}, back-propagates only the top $k$ most confident predictions within each weakly-labeled mini-batch. We set $k$ to $\nicefrac{1}{4}$ of the mini-batch size.
 
\item \textbf{Self-Weighted Weak Supervision (SW-WS).} Our proposed approach is described in Algorithm~\ref{trainingalg}. The classifier can be trained using exclusively weak labels ($W\!-$only) or both weak and strong labels ($W \! \cup \! S$). In either case, the weak supervision signal is weighted by self-computed confidence scores, corresponding to the predicted probability for the correct (weak) label. Therefore, image patches that are not characterized by the overall Gleason score label assigned to their respective WSI may contribute less to the gradient updates.
\end{itemize}

\begin{algorithm}
\caption{Training algorithm combining weak and strong labels}\label{trainingalg}
\begin{algorithmic}[1]

\State \textbf{Input:} model $\mathcal{M}$ with parameters \bm{$\theta$}, strongly-labeled dataset $S$, weakly-labeled dataset $W$, learning rate $\eta$, number of classes $K$.
\For{each training iteration $t$}
\State Sample data batches $ b_s = (x_s, y_s) \sim S, \; b_w = (x_w, y_w) \sim W $
\State Compute model predictions $ \hat{y}_s = \mathcal{M}(x_s; \bm{\theta}_{2t}) $
\State Back-propagate strong labels $ \bm{\theta}_{2t+1} \leftarrow \bm{\theta}_{2t} - \frac{\eta}{|b_s|} \sum\limits_{i=1}^{|b_s|} \nabla_{\bm{\theta}_{2t}} \mathcal{L}(y_s^{(i)}, \hat{y}_s^{(i)}) $
\State Compute model predictions $ \hat{y}_w = \mathcal{M}(x_w; \bm{\theta}_{2t+1}) $
\State Compute per-example confidence scores $ c_i = \sum\limits_{k=1}^{K} \hat{y}_w^{(i)(k)} \mathcal{I}\{ y_w^{(i)} == k\} $
\State Back-propagate weak labels $\bm{\theta}_{2t+2} \leftarrow \bm{\theta}_{2t+1} - \frac{\eta}{|b_w|} \sum\limits_{i=1}^{|b_w|} c_i \nabla_{\bm{\theta}_{2t+1}} \mathcal{L}(y_w^{(i)}, \hat{y}_w^{(i)}) $
\EndFor
\end{algorithmic}
\end{algorithm}

\section{Results}
 
 \textbf{Data preprocessing.} Both datasets (Table~\ref{table_dataset}) comprised formalin-fixed paraffin-embedded (FFPE) tissue samples, stained with H\&E and digitized at $40\!\times$/$20\!\times$ resolution. We extracted image patches of size $400\times400 $ at $20\!\times$ resolution and downsized them to $224\times224$. For the TMA data, we used all patches from the annotated ROIs. For the WSIs, we used the Blue Ratio transform~\cite{Del_Toro2017-gk} to prioritize patches with high concentration of cell nuclei and extracted the top 2000 patches per image.

\begin{table}[ht]
  \vspace{-2.5mm}
  \caption{Summary statistics of the TCGA and TMA datasets.}
  \label{table_dataset}
      \centering
      \begin{tabular}{lcccccccc}
        \toprule
        dataset & $\#$ patches & $\#$ cases & Gleason low/high & $\leq$6 & 7=3+4 & 7=4+3 & 8 & 9-10  \\
        \midrule
        TCGA &  $\sim$ 300'000  & 447 & 261/186 & 44 & 125 & 92 & 65 & 121  \\
        TMA  &  $\sim$ 25'000   & 886 & 524/362 & 403 & \multicolumn{2}{c}{121} & 226 & 136   \\
        \bottomrule
      \end{tabular}
\end{table}

\textbf{Model training \& evaluation.} In all experiments, we used CNNs with ResNet18~\cite{He2016-zi} architecture, categorical cross-entropy loss, Adam~\cite{Kingma2014-ff} optimization with base learning rate $1e\!-\!4$ and batch sizes 32/128 for the TMA/TCGA data, respectively. Model performance was evaluated via 5-fold cross validation. Within each CV fold, $20\%$ of the training TCGA data was held out and used for early stopping. For testing, a final score was derived for each WSI as the ratio of predicted high-grade patches over all patches. We report ROC AUC and accuracy for the binary classification task, as well as the Kendall's $\tau$ correlation coefficient between the WSI rankings produced by (a) the predicted scores and (b) the Gleason score groups (6, 7=3+4, 7=4+3, 8, 9-10) used in clinical practice~\cite{Epstein2016-cj}.
  
In the covariate shift reduction benchmark (Table~\ref{table_AUC1}), we observed that the color-jittering and stain transfer data augmentations improved model predictions on the target domain, whereas additional domain adaptation constraints did not have a big impact. Thus, we adopted the stain transfer and color-jittering augmentations in our subsequent data integration experiments.  

\begin{table}[ht] \centering
	\vspace{-2.5mm}
  \caption{Results of the covariate shift reduction benchmark on TCGA WSIs (5-fold CV).}
  \label{table_AUC1}

      \begin{tabular}{lcccccc}
        \toprule

        \rule{0pt}{3ex} & \multicolumn{2}{c}{AUC (stdev)} & \multicolumn{2}{c}{accuracy (stdev)} & \multicolumn{2}{c}{Kendall's $\tau$ (stdev)} \\
        \cmidrule(l){2-3} \cmidrule(l){4-5} \cmidrule(l){6-7}
        w/o color augm.     &  0.738 & ($\pm$ 0.062)  & 0.682 & ($\pm$ 0.049)  & 0.370 & ($\pm$ 0.078)  \\
        w/ color augm.      &  0.771 & ($\pm$ 0.042)  & 0.721 & ($\pm$ 0.032)  & 0.408 & ($\pm$ 0.041)  \\
        stain transfer      &  0.799 & ($\pm$ 0.034)  & 0.734 & ($\pm$ 0.045)  & 0.445 & ($\pm$ 0.049) \\

        \cmidrule(l){1-1} \cmidrule(l){2-3} \cmidrule(l){4-5} \cmidrule(l){6-7}
        MMD          &  0.786 & ($\pm$ 0.037)  & 0.723 & ($\pm$ 0.043) & 0.434 & ($\pm$ 0.046) \\
        CORAL        &  0.797 & ($\pm$ 0.030)  & 0.741 & ($\pm$ 0.033) & 0.438 & ($\pm$ 0.030) \\
        adversarial  &  0.802 & ($\pm$ 0.032)  & 0.738 & ($\pm$ 0.023) & 0.447 & ($\pm$ 0.044) \\

        \bottomrule
      \end{tabular}

\end{table}

In the data integration experiments (Table~\ref{table_AUC2}), we observed that the classifiers utilizing TMA labels performed better than the ones trained exclusively on TCGA weak labels. Furthermore, adding self-weighted weak supervision to the jointly-trained classifier resulted in the best overall performance.

\begin{table}[ht]
	\vspace{-2.5mm} 
  \centering
  \caption{Results of the data integration benchmark on TCGA WSIs (5-fold CV).}
  \label{table_AUC2}
      \begin{tabular}{lcccccc}
        \toprule
        \rule{0pt}{3ex} & \multicolumn{2}{c}{AUC (stdev)} & \multicolumn{2}{c}{accuracy (stdev)} & \multicolumn{2}{c}{Kendall's $\tau$ (stdev)} \\
           \cmidrule(l){2-3} \cmidrule(l){4-5} \cmidrule(l){6-7}
        $W\!-$only            &  0.814 & ($\pm$ 0.057)  & 0.754 & ($\pm$ 0.042)  & 0.445  & ($\pm$ 0.071)  \\
        $W\!-$only (MIL-WS)   &  0.815 & ($\pm$ 0.030)  & 0.756 & ($\pm$ 0.032)  & 0.446  & ($\pm$ 0.038)  \\
        $W\!-$only (SW-WS)     &  0.778 & ($\pm$ 0.055)  & 0.734 & ($\pm$ 0.047) & 0.385  & ($\pm$ 0.097)  \\
        $W \! \cup \! S$          &  0.845 & ($\pm$ 0.065)  & 0.799 & ($\pm$ 0.051)  & 0.510  & ($\pm$ 0.072)  \\
		$W \! \cup \! S$ (MIL-WS) &  0.839 & ($\pm$ 0.020)  & 0.792 & ($\pm$ 0.011)  & 0.504  & ($\pm$ 0.027)  \\
        $W \! \cup \! S$ (SW-WS)   &  \textbf{0.882} & \textbf{($\pm$ 0.024)} & \textbf{0.848} & \textbf{($\pm$ 0.010)} & \textbf{0.540} & \textbf{($\pm$ 0.032)}  \\
        \bottomrule
      \end{tabular}
  \vspace{-2mm}
\end{table}

\section{Discussion}
We have presented an approach that efficiently leverages weak and strong supervision signal for histopathology image classification. We believe that other medical image analysis tasks are likely to benefit from similar approaches, as medical expert annotations at the ROI-level are challenging to obtain, whereas patient-level information is often readily available.

\printbibliography

\end{document}